\pdfoutput=1

\documentclass[11pt]{article}

\usepackage[final]{emnlp2021}

\usepackage{times}
\usepackage{latexsym}

\usepackage{hyperref}
\usepackage{graphicx}

\usepackage{url}
\usepackage{xurl}

\newcommand{\RomanNumeralCaps}[1]
    {\MakeUppercase{\romannumeral #1}}
    
\usepackage[T1]{fontenc}

\usepackage[utf8]{inputenc}

\usepackage{microtype}

\title{Refocusing on Relevance: Personalization in NLG}

\author{Shiran Dudy \\
  Department of\\  Computer Science \\
  University of Colorado \\
  \texttt{shdu9019@colorado.edu} \\\And
  Steven Bedrick \\
  Department of \\Medical Informatics and\\ Clinical Epidemiology \\
  Oregon Health \& \\ Science University \\
  \texttt{bedricks@ohsu.edu} \\\And
    Bonnie Webber \\
  Institute for\\ Language, Cognition \\and Computation \\
  School of Informatics \\
  University of Edinburgh \\
  \texttt{bonnie.webber@ed.ac.uk}}

\begin{document}
\maketitle
\begin{abstract}


Many NLG tasks such as summarization, dialogue response, or open domain question answering focus primarily on a source text in order to generate a target response.
This standard approach falls short, however, when a user's intent or context of work is not easily recoverable based solely on that source text-- a scenario that we argue is more of the rule than the exception.
In this work, we argue that NLG systems in general should place a much higher level of emphasis on making use of additional context, and suggest that \textit{relevance} (as used in Information Retrieval) be thought of as a crucial tool for designing user-oriented text-generating tasks.
We further discuss possible harms and hazards around such personalization, and argue that value-sensitive design represents a crucial path forward through these challenges.

\end{abstract}

\section{Introduction}

The evaluation of natural language generation tasks (such as automatic summarization, machine translation, and dialogue generation) is commonly framed as one of comparing an automated system's generated output to some reference output,\footnote{In some cases, a 
small set of reference outputs may be used, as in BLEU's original formulation. 
} with the goal of achieving as close an alignment as possible.
Implicit in this experimental framing is the notion that for any given system input, there must exist a single, ``correct'' output. 

While this may arguably be a necessary simplifying assumption in terms of experimental evaluation,\footnote{
Statistical metrics based on this assumption often fail to reflect human judgments of quality or performance; 
see
\citeauthor{novikova-etal-2017-need}'s recent analysis.
} this ``one-size-fits-all'' philosophy has also constrained the ways in which NLG systems are designed, trained, and deployed.
For example, 
standard approaches to automated document summarization and machine translation rely only on the original source text, and do not typically take into consideration contextual factors involving the user or their situation of use. 
The problem is even larger when we consider NLG systems whose outputs are less firmly grounded to some specific input text, such as dialogue and question-answering systems.

This has a number of negative effects on NLG system performance and utility.
One such effect is system output that does not meet a given user's needs: for example, a summarization system that chooses to focus on different aspects of the source text than those in which the user was interested, or a machine translation system whose output is of an inappropriate register of formality.
Other effects are more subtle; for example, consider that, by designing systems to produce a single universal output, we amplify the effects of label and sample bias in training data~\citep{shah-etal-2020-predictive}, since when there is only one ``right'' answer, a statistical model will generally tend towards whatever it has seen most frequently during training.
One example of this is the well-documented behavior of machine translation systems defaulting to ``male'' for certain categories of phrase in gender-marked languages~\citep{Prates:2019bj}; another can be seen in the tendency for neural language models to over-predict frequent words~\citep{dudy-bedrick-2020-words}.

Recent work by \citet{flek-2020-returning} made a compelling argument for an increased emphasis on user- and task-level personalization in NLP applications, particularly those involved in classification or prediction. 
In this work, we build on their foundation and turn our attention specifically to tasks involving natural language generation.
While the paramount importance of user-level personalization in NLG applications has long been known in our field~\citep{rich_1979, kass1988modeling}, recent years have seen 
reduced
emphasis on this aspect of system design and evaluation.
We challenge the universalist simplifying assumption that underlies much of how such systems are built and evaluated today, and discuss how concepts from information retrieval --- specifically different notions of \emph{relevance} --- may be fruitful tools for considering personalization in the context of common NLG tasks.
We discuss several case studies in which an NLG system would benefit (or indeed, require) additional context about the user and their task, and discuss some possible challenges and harms that could be introduced by textual personalization in NLG.
Finally, we discuss how design methodologies that center the needs, goals, and rights of users may provide a crucial path forward to developing more robust and personalized NLG systems; specifically, we point to Value-Sensitive  Design~\citep{friedman2019value} as one such option.

\subsection{The Universal Truth Assumption}

The notion of a ground truth in which each annotated instance in a dataset corresponds to a single ``right'' answer has been already challenged in the work of~\citet{aroyo2013crowd,aroyo2015truth} in the context of language processing, where the authors proposed that instead of the ground truth assumption of a single correct target to a particular task, the crowd truth approach assumption generalizes more optimally as they reflect a set of various perspectives and interpretations, rather than one based on a single answer. 
\citeauthor{aroyo2013crowd} argued that in many NLP tasks, low inter-agreement among annotators is expected due to relatively ambiguous instances (sources) which subsequently triggers 
annotators in different ways, explaining the multiple perspectives (targets) collected. 

We argue there is room for more than a single strategy to overcome the ambiguity of the source, so while \citeauthor{aroyo2013crowd} proposed to embrace the diverse set of responses resulted from ambiguous input, we propose to further specify the under-specified source in order to generate more relevant outcomes to a particular user or scenario.
Particularly, we consider scenarios in which 
the source may be specified by adding indirect information about the user and their situation, and thereby contribute to reducing the ambiguity to a more limited set of predictions 
so 
that, given the same source, the target output may vary depending on the additional meta-data provided.


Following that, one may ask \emph{what} indirect and contextual information would be appropriate for use in textual personalization of NLG systems. 
Necessarily, there is no single answer to this, as it is entirely task- and user- dependent.
Similarly, when we consider the question of \emph{evaluation} of NLG systems, we also find a similar situation: there is no single universal ``right'' way to evaluate an NLG system (in the way that, say, variations on classification accuracy are generally appropriate for most classification problems).
Even ``zooming in'' on a specific NLG task --- abstractive summarization, for example --- notions of what makes a ``good'' summary quickly become worryingly fuzzy,\footnote{See section~\ref{how_to_benefit} for further discussion on this point.} and the situation only becomes more so when we begin to account for issues of user-level personalization. 

To move forward, we draw inspiration from the field of information retrieval, which has long faced a similar quandary in its decades-long quest for agreement on the notion of ``relevance.''

\subsection{Relevance: A Path Forward}
\label{sub:relevance_a_path_forward}

The key purpose of information retrieval systems has historically been framed as one of providing users with documents (or other information objects, such as images, etc.) that are ``relevant'' with respect to an information need. 
The question of how best to define what is meant by ``relevant'' has dogged the field since its inception (see \citet{saracevic_1975} for a classic early review of the topic\footnote{
See \citet{Saraevic:2006p9216,saracevic_2007} 
for more recent
reviews.}),
and has been the focus of a great deal of research from many different perspectives. 
\citet{borland_2003} grouped this body of work into two broad families of models of relevance. 
The first, and oldest, frames relevance as a question of topicality: a search result was relevant if it addressed a topic included in a search query. 
This model has certain similarities to the way that we often think of
NLG tasks: the system's behavior is measured in terms of the relationship between a given input and a given output, compared in a fairly constrained way.\footnote{In
IR,
the question is whether
or not 
a concept from the query is mentioned in the document ; where analogously in NLG we use
overlap-based metrics such as BLEU or ROUGE.}
The user is absent from the conversation, as are specifics of what they are trying to accomplish. 
This model is sometimes framed as a ``system-oriented'' or ``objective'' model of relevance.

This model is not without its uses, but has important limitations. 
As a simple illustration of this, consider that a given search result may be topically aligned with some aspect of a search query... but if it covers information our user already knows, can it truly count as being ``relevant?'' 
The second family of relevance models, labeled by \citeauthor{borland_2003} and others as ``user-oriented'' or ``subjective'' relevance, addresses these limitations by conceiving of ``relevance'' as a more complex, dynamic, and multidimensional construct. 
Crucially, this view of relevance necessitates the consideration of who, exactly, we are imagining as our user, and what it is that they are attempting to accomplish.

There are many different models of user-oriented relevance, but for the purposes of considering NLG applications, we find \emph{situational relevance}~\citep{schamber_1990} to be particularly useful.
Under this model, relevance is understood as ``the utility or \emph{usefulness} of the ... information objects'' in terms of  ``the relationship between such ... objects and the work task at hand underlying the information need as perceived by the user''~\citep[p. 915, emphasis original]{borland_2003}.
\footnote{\emph{Situational relevance} as introduced by \citeauthor{schamber_1990} is meant as a broad and somewhat abstract term,  and encompasses to the totality of the situation surrounding the use of a tool, including the user, their setting, and their task. Elsewhere, the word ``situation'' is sometimes used in a more narrow and concrete manner, referring to either the setting in which a task takes place or to a specific task itself. }
Another
related
notion of relevance that is highly applicable to NLG is that of task-oriented relevance, in which the notion of relevance regarding a system output is explicitly tied to that output's impact on the user's ability to complete their task. 
As 
defined by \citet{hjorland2002work}, ``Something (A) is relevant to a task (T) if it increases the likelihood of accomplishing the goal (G), which is implied by T.''

While there are many types of NLG applications, with many different objectives, all of them share the underlying goal of communicating a given text/utterance to a human user in helpful and relevant way. 
Looking through the lens of 
situational and task-oriented relevance, 
we posit that 
for any such application, the notion of there being a single ``correct'' output for a given input is nonsensical.
Adopting a one-size-fits-all approach to system development and evaluation is thus more than a simplifying assumption; it ignores a key 
and integral 
aspect of system behavior.

Only by taking into account the user and their task can the output of a system be made relevant; one way to achieve that is through communicating information to a user in ways that are relevant to them and/or their situation. Early work on personalized natural language systems such as that of~\citet{kass1988modeling} recognized this, and placed heavy emphasis on building systems around rich user models. More recently, work by \citet{newman-etal-2020-communication} makes the link directly to the core purpose of an NLG system, framing the problem as one of modeling the ``communicative function of language.''
They point out that ``a speaker's goal is not only to produce well-formed expressions, but to convey relevant information to a listener''; in the context of NLG, this must necessarily take a personalized form.

As an example of how a user's immediate situation may relate to the content and pragmatics of generated language, consider a user who asks a question-answering system ``how to put out a fire.'' 
If they are in the kitchen standing over a grease fire, a situationally-relevant response would be short in length, very focused on a specific firefighting technique, and delivered in an imperative register. 
In a different situation (perhaps while sitting around the dinner table with inquisitive children), a  situationally-relevant response to the same query might include more generic informational content about firefighting, and be delivered in a more narrative mode.


Following this example, we now turn
discuss
several different NLG application areas, and the role
personalization may play in each.

\section{A Survey of Textual Personalization}
    
\subsection{The Status Quo}~\label{status_quo}

The goal of textual personalization in NLG is to develop models that generate \textit{relevant} text to users in response to information (or other) needs, expressed in natural language, the exact form of which will depend on the specific application. 
Asking a question, or interacting in a dialogue, does not happen in a void; rather, there is an underlying intent for doing so. 
The problem is that current text generation tasks operate independently from the users they aim to support, which subsequently limits their usefulness when deployed. 
What are the limitations introduced by not modeling users in the system?

In abstractive summarization the dominant assumption is generally that there is a single type of summary that can be produced from a given text.
This applies also to evaluation scenarios that make use of multiple reference outputs, as in such cases an implicit assumption is that the references should typically exhibit minor linguistic variation, rather than summaries that vary substantially in their contents~\citep{cachola2020tldr,see2017get,grusky2018newsroom,harman2004effects}.
While this approach simplifies the development and evaluation process, in practice, different users would likely find different aspects of the source article to be more relevant to their needs than others; in other words, if an original article includes facts $A$ through $E$, one user's optimal summary might involve facts ${\{A,B,C\}}$ while another user's would instead feature ${\{A,C,D\}}$. This example will be referred as the source-to-target transformation example. 
Beyond the factual content of a summary, users could also vary in terms of the level of detail that they would find useful~\citep{louis-nenkova-2011-text}, 
some would enjoy in-depth summarization, while other users
would benefit from a simplified writing style~\citep{scarton2018text}. Similarly, in a paraphrasing~\citep{witteveen2019paraphrasing} task, two different users, each with different intents and goals, would likely find different paraphrases to be ``correct.'' 
A development and evaluation paradigm which assumes a single reference output (or a small set of semantically-equivalent reference outputs), is unlikely to support (or encourage) the generation of user- and task-personalized output. 

Thus far, we have described families of task that we will henceforth refer to as \textit{source-to-target transformation} generation tasks.
This category encompasses tasks that are firmly grounded in a specific source text, and which must produce fluent textual output whose content is similarly grounded to that same source, e.g. automated summarization, paraphrasing, narrative report generation, etc. 
We will now discuss a second category of generation, which we refer to as \textit{query-response}.
and includes question-answering, dialogue agents, and the like.
Query-response tasks are generally guided by a query or prompt from a user (or by a series of such prompts), and while they may draw on source documents of various sorts to inform their outputs, the link is much less straightforward than in the transformation tasks that we have previously discussed.
User-level personalization, however, is perhaps even more essential to query-response tasks.
Consider, for example, open domain question answering: for any given question, there might be multiple factually-correct answers~\citep{yang2015wikiqa}, but without modeling a user in this dynamic, it is difficult to say whether any particular answer will or will not be be relevant. 

Open domain dialogue systems~\citep{li2016deep,wensemantically} face a somewhat related problem to what is described above from a technical standpoint, in that there is no one ground truth that is expected.
Dialogue agents tend to generate general prompts that may address a given question (the query) and present a natural and informative response, but are indifferent to the user, irrespective of anything but the question it was asked about. 
Crucially, however, there has
been significant attention paid to the problem of personalizing dialogue agents, to a much greater degree than is the case in other realms of NLG.

Some of this work has focused on the use of psycho-linguistically informed parameters (verbosity, etc.) to tune a system's output~\citep{mairesse-walker-2011-controlling}, while other work has focused on ways to make use of more general ``personas,'' meant to represent salient features of the dialogue agent or its interlocutor (or both), and to use those features to influence the semantic and stylistic content produced by the agent~\citep{li-etal-2016-persona,zhang-etal-2018-personalizing}.
Going the other direction, \citet{madotto-etal-2019-personalizing} and others have worked to infer relevant properties of the interlocutor from the conversation itself, rather than relying on a pre-specified persona.
An additional direction of work in dialogue personalization has been in efforts to have automated dialogue agents behave ``empathetically'' with their interlocutors, by attempting to match the register and contents of their output with what they perceive to be the emotional state of their user~\citep[see e.g.][]{lin-etal-2019-moel}.\footnote{
These directions can be combined 
~\citep{zhong-etal-2020-towards}.} 

It is perhaps unsurprising that dialogue systems have focused on personalization to a larger extent than have other types of NLG application, as dialogue systems are deeply and necessarily user-oriented, both in terms of their design and their evaluation, in a way that other types of NLG are not traditionally thought of as being. 
In human-human interaction we communicate in a more collaborative fashion, considering what may be additional information required to solve a problem, knowing that the same question might be responded differently depending on the user's age, situation, gender, expertise, register, patience, and underlying intent when posing a question.

Thus, when a dialogue system \emph{fails} to perform in this way, it represents an obvious failure of the system, much more than slightly agrammatical output in a generated summary might, for example.
Put in terms of the ``fluency'' and ``adequacy'' dimensions often used in MT evaluation~\citep{white-etal-1994-arpa}, the two are much more closely tied together in the case of a dialogue system than they are in an MT system.
All of that said, current neural-network-based dialogue systems are very much in their infancy with regards to personalization; in a recent work discussing aspects of human-computer interaction in relation to dialogue systems, \citet{kopp2021revisiting}  suggest that the field should ``(re-)emphasize the hallmarks of human communication and its complexity, and ... argue that we should not lose sight of these hallmarks when deriving requirements for human-agent-interaction.'' 
In our work, we are particularly interested in shedding light on the end user and their needs when using textual communication systems. 


\subsection{Benefits of Personalization in NLG}
\label{how_to_benefit}

Different NLG tasks can benefit from personalization in different ways.
For example, the utility of an automatically-generated document summary will depend heavily on the context of use, as noted by~\citet{sparck1998automatic}, who listed several factors that come into play in the task of human summarization, specifically in terms of a \textit{purpose} factor that considers the situation, audience, and use -- all of which are context dependent. 
As an example, consider a scientific article that describes a dataset, a model architecture, and an evaluation methodology.
A summarization method that takes into account the needs and interests of its user may be able to tailor its output to focus on the most contextually-relevant aspect of the article.
Similarly, paraphrasing systems could benefit from a notion of situational relevance, as different users will need different aspects of the source document to be retained, depending on their scenario of use. 

Context is also critical for open-domain question-answering, as many questions are  under-specified when considered solely in terms of their textual contents.
Consider the question ``how to drill a hole in the wall?''
Without knowing more about the tools available, the composition of the wall, etc., this question is simply unanswerable; a one-size-fits-all attempt is likely to be irrelevant at best.

In terms of \textit{dialogue response}, \citet{adiwardana2020towards} presented a system that allows for diverse responses to be elicited by the same prompt, and \citet{tevet2020evaluating} proposed steps towards measuring content diversity of responses, which is aligned with the crowd truth proposal mentioned earlier~\citep{aroyo2013crowd}. 
While this diversity should be encouraged and expected, the next step is to learn how to generate these responses in a more \textit{controllable} way that could enable us to reason why the particular target was generated, instead of consenting to randomly plausible response. 
One way to get this control is by constructing responses that are relevant in a particular context, either providing them as additional information on the user or situation, or training agents to identify and resolve knowledge ambiguities through clarifying questions in order to respond to a user in a relevant way.

In addition, perceiving the diverse responses to be a result of \textit{additional factors} other than the source alone contributes to a more coherent and explainable outcome enabling measurement of the effect of the additional data on the final target.     

For instance, in a scenario where a customer speaks to a hospital representative, the same question might elicit one response if the caller is a patient, and a very different kind of response (in terms of both content and register) if the caller is a medical provider.
So while there may be various plausible/correct and diverse responses generated by the system, some of them are mutually exclusive and are dependent on the additional data provided.
Furthermore, in this scenario, we note that some possible responses may bear a risk of compromising sensitive data. 
As another example, a question that a child raises to a voice assistant such as Alexa is expected to be answered differently than when raised by an adult, often in a simplified fashion as described by~\citet{scarton2018text}; this task may also involve revising both the content and the syntax of the response.

\section{Potential Harms from Textual Personalization}\label{harms}

Potential harms from personalization can be grouped into several categories.
The first stems from the fact that personalized information systems must necessarily have access to and make use of personal information about their users, which brings with it the risk of that private data being improperly disclosed or otherwise misused~\citep{krishnamurthy2011privacy,corrigan2014does}. 
A second, related category of potential harm is that of such an information system being used by its operators to leverage its capabilities against the interests of its users; possible scenarios include targeted advertisements for predatory educational or financial products being shown to vulnerable individuals, or a system that shows or hides certain job listings based on the gender of its user (or, in truth, based on the system's limited model thereof).

An additional, more subtle family of risk has to do with the indirect consequences of widespread deployment of personalized information systems. 
In social media, and search engines, the choices made by the model developer when incorporating the users' data are based on the confirmation bias theory proposed by \citet{nickerson1998confirmation}, describing the tendency to favor information that confirms one's prior beliefs.~\footnote{As opposed to vulnerabilities related to a specific user, confirmation bias susceptibility is across users and therefore is mentioned separately.}
The assumption is that presenting a user with evidence that supports their belief would increase their engagement, resulting in higher margins for these companies. 
Strategizing around this goal was shown to contribute to the formation of `filter bubbles'/`echo chambers' decreasing the users' exposure to diversity of perspectives \citep{flaxman2016filter,chitra2020analyzing} 
as well as radicalization~\citep{maddox2020interrogating}. 
This intellectual isolation can erode the healthy functioning of democracies, which thrives on exposure to multiple points of view~\citep{downs1957economic,baron1994electoral,lassen2005effect}.

To illustrate this, \citet{golebiewski2018data} describe the phenomenon of searching on ``fragmented concepts,'' where politically-involved queries return only results that support the query's narrative instead of exposing the user to both sides. 
\citeauthor{golebiewski2018data} argue that intentionally avoiding returning the results containing the other side stems from the fear of having the search engine portrayed as politically biased. 
Users are often unaware of the active role that personalized software is taking in this process; \citet{tripodi2018searching} describes an interview with a participant who ``admitted that her Google searches rarely revealed alternative points of view. However, she did not consider how her returns were tied to her own search practices or Google’s algorithmic ordering of information.''
\citeauthor{tripodi2018searching}'s interviews reveal how users ascribe the lack of alternative views to their nonexistence, rather than search manipulation. 

Finally, another possible concern in the process of personalizing a system is that the system's assumptions (and thus its personalized model of the user) may be incorrect, resulting in output that is irrelevant at best, and misleading at worst.
For instance, a recipe system supporting a user in the task of making enchiladas might instruct its user to ``fold in the cheese''; depending on the user's level of experience as a cook, this may be an unfamiliar usage of the word ``fold,'' and it may be more appropriate to instruct them to ``gently incorporate the cheese with a scooping-and-folding motion''~\citep{Christensen:2013aa}.
Similar issues could arise in other contexts, with more serious results; imagine a system providing personalized medical advice, for example.

\section{Value-Sensitive Design}~\label{value}

To design a system that can address the shortcomings described in Section~\ref{harms}, we can follow research in value-sensitive design ~\citep[VSD,][]{friedman2019value} that is aimed at identifying the stakeholders and prioritize the values to instill in the system under the assumption that ``in designing tools we are designing ways of being.'' 
How might this affect the direct or indirect stakeholders of the system? 
In this section, we focus on the goals of the user of the system. 
There is of course no single answer to what this looks like, but in this section we attempt to dissect the problem in order to evaluate the possible outcomes of personalization through the lens offered through VSD. 

One value achieved by personalization is that of \textbf{accessibility} of information. 
A summary written for a child may be realized differently than one for an adult. 
By learning about the user, the system can adjust the syntax/register/style it produces, as well as the content, to make it more comprehensible to the user~\citep{scarton-etal-2018-text}. 
However, this requires that \textbf{privacy} also be taken into account, as personalized systems make use of data about their user. Since these values may be at odds with each other, understanding their possible trade-offs is necessary.
By cautiously balancing the most relevant data the system requires for personalization, as well as the privacy concern of its users, system designers may optimize the amount and types of data that the user may be required to turn in.

Another indirect outcome of accessibility of information can be shown in discourse generation, where, for example, the quality of the \textbf{interaction} can elicit user responses that can more easily (or more quickly, or more accurately) direct the system towards the user's goals of the conversation. 
In other words, an accurate and thorough (language) understanding of the user's response may help build a shared ground and ease the transfer of communicating ideas, intents, or requests which subsequently accomplish the user's goal in a conversation. 

\textbf{Safety} can be another important value to be accountable for as developing models that interact with humans. ~\citet{dinan2021anticipating} elaborate how a dialogue agent can cause real world harm by proactively introducing harmful language or content,\footnote{What \citeauthor{dinan2021anticipating} refer to as "instigating."} or posing as an expert, which can mislead and create harm in the physical world (providing medical advice for {COVID}-19, for instance). Of course, source-to-target transformation tasks are susceptible to similar problems, as well.

Another opportunity such systems bear is that of
enabling a proactive personalization that can provide a sense of \textbf{agency} to a user. This can be done through allowing them to define their specific needs and interests, and by that providing a sense of control over the system as described by~\citet{synofzik2008beyond}. 
A personalized system can find ways to reason/explain on the type of data presented to its users, promoting \textbf{transparency}. For instance to paraphrase a sentence for a user, a system can indicate what were the relevant dimensions it maintained and why (see the source-to-target example in section~\ref{status_quo}).
Moreover, if a user knows how their data is used that may contribute to developing \textbf{trust}. \citet{ribeiro2016should} identified two types of trust; the first, whether a user trusts an individual \textit{prediction} sufficiently to
take some action based on it, and the second, whether the user trusts a \textit{model} to behave in reasonable ways if deployed ``in the wild.'' 
We add to that a third sense of trust, can we trust the \textit{developer}, that their knowledge integrated into these systems is serving the needs of the user faithfully and respectfully. Projecting aspects of value-sensitive design into text personalization tasks is the first step towards making a human-centered communication system. 

\section{Architectures for Personalization}

The previous section described the conceptual guidelines we propose, while this section is focused on practical aspects on how to implement a language generation system that allows for textual personalization to a situation or a user. 
At the very basic level, personalization of text can be done through keywords, much as a search engine attempts to return relevant text to a user in response to a query. 
However, keyword retrieval may not always recover the user's intent and as a result, the returns may not always be relevant. 
For instance the concept of ``neural network'' is highly associated with a particular sense, but has another sense that is actively used in the field of neuroscience, making some matches irrelevant. 
In this work we claim that the advances of natural language generation should be incorporated to personalization in ways that could benefit the user beyond good search engine returns. 

Since many NLG tasks are based on language models, we propose architectures that can generate personalized text through the personalization of language models.
\citet{khalifa2021a} present in their work a language generating system that enables generating text based on pre-trained language models, that can both apply pointwise as well as distributional constraints, while deviating a little as possible from the original distribution. 
The essence of this work in the context of personalization is the ability to generate text in a \textit{controlled} fashion. The aspect of control is crucial for the development of future systems.
One could adapt a language model to a particular user through conditioning on all outputs to speak about topics relevant to the user (a pointwise constraint) and have $50\%$ of them to be related to topic discussed in the given query (a distributional constraint).~\footnote{The original example conditioned all predictions to relate to sports, and $50\%$ of them reflect female characters.} 
A possible concern, at this point, when employing this model has to do with the ability within the same model to adapt different distributional constraints quickly. 

Another direction that could overcome this limitation presented by~\citet{li2021prefix}, where the model is conditioned on a prefix intended to provide context to steer the generated text. 
One way to incorporate this model is to condition the textual generation on different users, however, compromising centralized models bears higher costs to overall users' privacy~\citep{zyskind2015decentralizing}; instead, a per-user model can be trained on user's personal data, perhaps remaining entirely on their device. 
Moreover, ideally, implementing personalized agents can be done through Federated Learning~\citet{li2020federated}, where the model updates are made locally and globally through maintaining the user's privacy.\footnote{
In this case, protecting privacy comes at the cost of increased complexity.} 
As an example of applying the prefix scenario (by~\citet{li2021prefix}) to personalization, providing a prefix of ``How does the email my mom sent relate to my sister?'', together with the email in question as the source, would then generate a personalised textual output. 

While the focus of this work is mainly on text, it is plausible to assume that the greater objective of such work is to improve and enhance these communication models through the process extending the shared knowledge both parties maintain for the purpose of problem solving. 
Therefore, personalization may likely be more effective through integrating modalities beyond text and should be a cross disciplinary effort. 
When considering architectural perspectives for textual personalization, the realization of the information can vary greatly both across different individuals and within the same individual. 
In their work, following a survey on communication preferences in adults, ~\citet{himmelsbach2015enabling} concluded the due to high variation, it is generally recommended to support several modalities of communication, showing that there is no one size fits all approach.

In a survey on multimodality in educational setting, ~\citet{walters2010toward} describes the importance of the impairment-specific approaches to accommodate individuals with specific disabilities and contributing to an inclusive environment; furthermore, \citeauthor{takayoshi2007multimodal} describes how ``the more channels students (...) have to select from when composing and exchanging meaning, the more resources they have at their disposal for being successful communicators.'' 
The optimal modality may also depend on the situation, and it is reasonable to consider an event in which a user may opt for more than one modality within the same task; in the example of a driver asking Siri ``how can one recognize Mount Saint Helens?'', while an image would be a highly informative modality, it would provide a less optimal output at that precise point in time.
Instead, in order to avoid removing the driver's sight from the road, a verbal description would be a less risky alternative. 
At times one modality may complement another as shown by the work of \citet{wang2016low} and \citet{zhu2018msmo}, in which the output of a summarization task was presented not only through text but through images; in~\citet{wang2016low} the output was also structured temporally on a timeline. 
\citet{liao2018knowledge} introduced a multimodal dialogue agent for fashion retail where the visual appearance of clothes and matching styles are crucial in understanding the user’s intention. 
Returning to an earlier example, multimodality can help overcome a system's limitations. 
Recall the example of a recipe system providing its beginner-level user with the (unhelpful) instruction to ``fold in the cheese;'' if the instruction was accompanied by a video demonstrating of the process, this could help the user not only accomplish their immediate goal but also to expand their capabilities for next time by learning new vocabulary.
%


\section{Conclusions: How to Begin?}

One major obstacle to the development of personalized NLG systems is that such systems often depend on access to sensitive and personal data from users, making large-scale data resources difficult (or impossible) to obtain and share across a community.  
While this is a major challenge, we can draw inspiration from related fields that must work with such data and have developed best practices around how to do so, such as the biomedical domain.
One possible model is that of the MIMIC-\RomanNumeralCaps{3}~\citep{mimic2016} dataset, which is made up of thoroughly deidentified electronic medical records from thousands of patients from Beth Israel Deaconess Medical Center in Boston, MA.
This dataset is freely available to community members to use in research, but access is closely managed. 
Before gaining access to the dataset, users are required to take an online human-subjects research training module, which also contains content on the protocols for working with the dataset. 
Users of MIMIC-\RomanNumeralCaps{3} must also sign a data use agreement under which  they (and, importantly, their institutions) legally agree to restrictions on the use and redistribution of the data.

This solution has proven adequate for the MIMIC-\RomanNumeralCaps{3} dataset; however, there exist datasets where even heavily controlled redistribution is not an option.
In such cases, \emph{data enclaves} can be effective tools for community research on large, private datasets~\citep{lane2010balancing}.
One example of such an enclave is the National COVID Cohort Collaborative (N3C) system~\citep{haendel2020national}, which is run by the National Institutes of Health and stores harmonized demographic and clinical data about millions of patients from participating health care organizations.
After obtaining approval by their local institutional review board (IRB) and executing a formal data use agreement, users who have been granted access are able to analyze and interact with the dataset in a secure cloud computing environment.
The environment is designed such that data may not be removed or exported, and the platform includes capabilities for machine learning, statistical analysis, and data visualization. A similar enclave was used during the  2021 CL-Psych Shared Task~\citep{macavaney-etal-2021-community} in an NLP context, in order to allow the research community to interact with sensitive data relating to mental health in an IRB-controlled manner.

A related approach has been used in information retrieval in situations where even limited and controlled access to the original data is not possible.
Under the ``Evaluation-as-a-Service'' paradigm~\citep{lin_2013_eaas,eggel_2018}, developers virtualize their systems, and send them to a secure computing environment where they are trained and evaluated on an entirely private dataset, with the results being shared among the community.
While more limiting than a data enclave approach, experience in IR has shown that this model is a feasible one for shared-task evaluation~\citep{hopfgartner_2018}.
Such a system could be constructed for the design and evaluation of NLG applications involving sensitive data about individuals.

It may also be possible to augment more traditional data sets to simulate personalized behavior without actually requiring sensitive data.
For example, in the context of a paraphrasing task, one might supplement the original passage to be paraphrased with a set of features or questions that the algorithm should use to ground its behavior; the same input passage could be repeated with different grounding questions, and the evaluation design could take into account the degree to which the system's output was responsive to the additional data.

Similarly, in a summarization context, one could augment an input document with different sets of questions representing different users' needs, and an evaluation could take into consideration how well the output summary addressed the stated questions.
Open domain questions-answering datasets and tasks could be similarly constructed. 
In dialogue systems, beyond the persona-based approach previously discussed, there is a long tradition of explicit user modeling \citep{biswas2010brief,walker2004generation,kass1988modeling} with many approaches
that could be drawn on in order to generate simulated needs or goals for such a system to meet.

Another useful avenue to explore is that of richer evaluation methodologies. 
Rather than restricting our analysis to the system's output, truly assessing whether an NLG system is producing relevant personalized output necessitates extrinsic, task-based evaluations involving users. 
Such methods have long been used in the evaluation of automated summarization systems~\citep{hand-1997-proposal,he_1999,Mani:2001uc,mckeown_2005} as well as NLG systems~\citep{mellish_1999,reiter-etal-2001-using,colineau-etal-2002-evaluation}, though we note that recent years have seen somewhat less of this sort of ecologically valid evaluation, and much more focus on statistical evaluation; the work of \citet{barker-etal-2016-whats} and \citet{newman-etal-2020-communication} represent examples of very welcome exceptions to this trend.

\section{Future work}
In this opinion paper we argue for the benefit of personalizing NLG tasks. 
We hope that through this work and others' we will continue to make steps towards personalized text, from developing relevant focused datasets through methods that would make text more accessible and of practical use for real users.   

\section{Acknowledgements}
We would like to thank the anonymous reviewers who provided very useful comments. 
This research was supported by the NSF National AI Institute for Student-AI Teaming (iSAT) under grant DRL 2019805, as well as the National Institute On Deafness And Other Communication Disorders of the National Institutes of Health under Award Number R01DC015999. The opinions expressed are those of the authors, and do not represent views of the NSF or NIH.
\bibliography{custom}
\bibliographystyle{acl_natbib}




\end{document}